\pdfoutput=1
\documentclass[11pt]{article}

\usepackage[]{acl} %review

\usepackage{times}
\usepackage{latexsym}
\usepackage[T1]{fontenc}
\usepackage[utf8]{inputenc}
\usepackage{microtype}

\usepackage{booktabs}
\usepackage{graphicx}
\usepackage{multirow}
\usepackage{url}
\usepackage{subfig}

\usepackage{amsmath}
\usepackage{bbm}

\DeclareMathOperator{\Var}{\mathrm{Var}}
\DeclareMathOperator{\Cov}{\mathrm{Cov}}
\newcommand{\vect}[1]{\mathbf{#1}}

\title{Toward More Effective Human Evaluation for Machine Translation}

\author{Belén Saldías$^1$, George Foster$^2$, Markus Freitag$^2$, Qijun Tan$^2$ \\
  $^1$MIT Media Lab\\
  $^2$Google Research \\
  \texttt{belen@mit.edu}, 
  \texttt{\{fosterg, freitag, qijuntan\}@google.com} \\}

\begin{document}
\maketitle
\begin{abstract}
Improvements in text generation technologies such as machine translation have necessitated more costly and time-consuming human evaluation procedures to ensure an accurate signal. We investigate a simple way to reduce cost by reducing the number of text segments that must be annotated in order to accurately predict a score for a complete test set. Using a sampling approach, we demonstrate that information from document membership and automatic metrics can help improve estimates compared to a pure random sampling baseline. We achieve gains of up to 20\% in average absolute error by leveraging stratified sampling and control variates. Our techniques can improve estimates made from a fixed annotation budget, are easy to implement, and can be applied to any problem with structure similar to the one we study.
\end{abstract}

\section{Introduction}

As automatic natural language generation systems improve, evaluating them is becoming more challenging for both human and automatic methods \cite{elikyilmaz2020EvaluationOT,gehrmann2022repairing}.
In machine translation, this has led to increased adoption of techniques such as MQM \cite{freitag2021experts,freitag-etal-2021-results}, an elaborate error-based methodology for scoring output, typically carried out by skilled human annotators. 
While MQM is more accurate than traditional crowd-based Likert-type scoring, it can also be significantly slower and more expensive, creating a strong incentive to reduce annotation time and cost.

In this paper we investigate a simple solution to this problem, namely reducing the number of text segments that a human annotator must rate. We assume a basic scenario in which a single annotator is given a test set to rate, and the task is to predict the average MQM score they would assign to the whole set by having them rate only a selected subset. 
This is a natural and versatile way to deploy human annotation effort within a framework like MQM; it
differs from the tasks considered by recent work with similar motivation, which focus on system ranking \cite{mendoncca2021online,thorleiksdottir2021dynamic} or combining human and metric scores without the express aim of predicting human performance \cite{hashimoto2019huse,singla2021using}.
Although our experiments are carried out with MQM-based scores, our methodology is applicable to any setting in which numerical scores are assigned to items for later averaging.

We approach the task of choosing segments as a sampling problem, and investigate classical methods for reducing sample variance and bounding estimation error. To improve accuracy, we employ two sources of supplementary information. First, in keeping with recent practice, we assume segments are grouped into documents. This lets us exploit the tendency of segments within a document to be relatively homogeneous. Second, we make use of modern automatic metrics such as COMET \cite{rei-etal-2020-comet} and BLEURT \cite{sellam-etal-2020-bleurt} which correlate better at the segment level with human judgments than traditional surface-based metrics like BLEU \cite{papineni2002bleu}. These serve as a rough proxy for human scores.

We show that document and metric information can be used to reduce average estimation error by up to 20\% over a pure random sampling baseline. 
Due to high sample variance, it is difficult to reliably achieve a similar reduction in annotator effort for a given error tolerance.
However, we suggest an alternative perspective in which our technique can be used to improve estimates made on the basis of a fixed rating budget.
Although there is no guarantee of beating random sampling in any particular case, there is a high probability of improving on average. This improved estimator is easy to implement, and applicable to any human labeling task that produces numerical scores, and for which document membership and automatic metrics are available.

Our work is most similar to that of \citet{chaganty-etal-2018-price}, which we extend in several ways. We adopt their use of control variates, but consider multiple metrics rather than just one, including learned metric combinations; we also employ modern neural metrics rather than metrics based on surface information. % or context-free embeddings.
We combine control variates with stratified sampling using either proportional or optimal allocation, and additionally evaluate an incremental scenario in which sampling adapts to observed ratings. Finally, we investigate two analytical methods for bounding the error in our estimate.

\section{Methods}
\label{sec:methods}

We assume a fixed test set consisting of translated segment pairs, and a single human rater who assigns scores to segments. Each segment belongs to a document, and has an associated vector of scores from automatic metrics. Our goal is to select an informative subset of segments to be labeled by the rater, and use the subset to predict the average score that would have resulted if we had asked the rater to label the whole set. By exploiting document and metric information, we hope to reduce the number of segments that must be manually labeled.

Formally, let $x_1,\ldots x_N$ be the segment scores, $\mu = \sum_{i=1}^N x_i / N$ be the test-set score to be predicted, and $\sigma^2$ be the variance of the scores. 
The following side information is available for each segment $i$: an index $d_i$ that indicates its membership in one of $D$ documents, and a vector of automatic-metric scores $\vect{y_i}\in \mathbbm{R}^M$. Unlike the segment scores, which are only revealed if they are in the selected subset, the side information is always available for the whole test set. 

We approach this task as the problem of sampling $n \le N$ scores $X_1,\ldots,X_n$ without replacement from the test set and deriving an estimate $\hat{\mu}$ for $\mu$ from the sample such that $E(\hat{\mu}) = \mu$ (that is, $\hat{\mu}$ is unbiased) and $\Var(\hat{\mu})$ is as small as possible. Low-variance estimators make it more likely that the estimation error $|\mu - \hat{\mu}|$ will be small.
A baseline is to draw $n$ segments at random and compute their mean. This gives an estimate that is unbiased, with variance:
\[
\Var(\hat{\mu}) = \frac{\sigma^2}{n}\left(\frac{N-n}{N-1}\right)
\]
We investigated two classical unbiased strategies for reducing variance relative to this baseline:
stratified sampling
and control variates \cite{rice2007mathematical,bratley2012guide}.

\subsection{Stratified sampling}
\label{sec:sampling-methods}

\begin{figure}[t!]
\centering
\subfloat[Stratified sampling forces sampled segments (shown in red) to be evenly distributed across bins, resulting in better estimates when the score variance within bins is lower than the variance across bins.]{
  \includegraphics[width=\columnwidth]{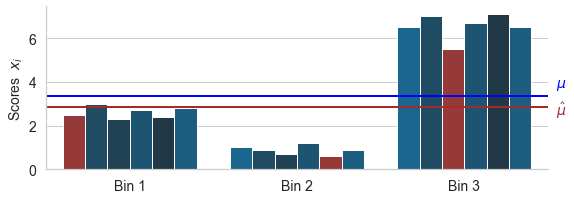}%
  \label{fig:stratified-sampling-diagram}
}

\subfloat[Control variates allow for reversing the shift of the sample mean $\bar{X}_n$ depending on the strength of the correlation between $X$ and $Z$. In this illustration, where $X$ and $Z$ are highly correlated ($\sim$0.9), $\bar{Z}_n < 0$ reflects the negative shift in $\bar{X}_n$.]{
  \includegraphics[width=\columnwidth]{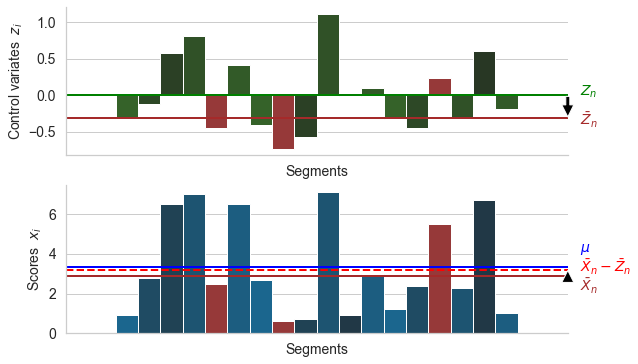}%
  \label{fig:control-variates-diagram}
}

\caption{Complementary strategies for reducing the variance of the estimated average score.}
\label{fig:sampling-diagram}
\end{figure}

Stratified sampling involves partitioning scores into bins that group similar items, then sampling some items from each bin. Intuitively, the idea is that if the variance within each bin is low, drawing too many samples from a particular bin is inefficient because it only serves to improve an already good estimate---therefore the sample should be spread evenly (in some sense) across bins. See Figure~\ref{fig:stratified-sampling-diagram} for an illustration. As a side benefit, having human scores more evenly distributed across different types of segments is a useful characteristic if the labeled segments are to be the subject of further analysis.

Formally, suppose the test set is divided into $L$ bins, where bin $l$ contains $N_l$ segments of which $n_l$ have been sampled, with sample mean $\hat{\mu}_l$. Then the stratified estimate is:
\begin{equation} \label{eqn:strat-sampling}
\hat{\mu} = \sum_{l=1}^L \hat{\mu}_l\, N_l / N.
\end{equation}
It is easy to verify that this is unbiased.

Stratified sampling requires a method for partitioning the test set into bins and a way of allocating the  $n$ segments in the sample to individual bins. We investigated two methods for partitioning the test set: by documents and by metric-score similarity. The optimal (lowest variance) allocation assigns segments proportional to a bin's size and variance:
\begin{equation} \label{eqn:alloc}
n_l = n\, \frac{\sigma_l N_l}{\sum_{l=1}^L \sigma_l N_l}.
\end{equation}
Since the bin variances $\sigma_l$ are unknown, a conservative strategy is to assume they are all equal, resulting in pure proportional allocation: $n_l = n\, N_l / N$.
A potential enhancement is to approximate optimal allocation using estimated variances
$\hat{\sigma}_l \approx \sigma_l$ derived from the metric scores in each bin.

Two technical issues arise in stratified sampling. First, the per-bin sizes specified by equation (\ref{eqn:alloc}) are not necessarily whole numbers. This can be solved using a rounding scheme that minimizes $\sum_{l=1}^L |n_l - n_l'|$, where $n_l'$ are whole numbers that sum to $n$.
A second problem is that $n_l$ can be greater than the number of available segments $N_l$
when using optimal allocation in high-variance bins. When this occurs, we choose the bin for which $n_l - N_l$ is largest, set $n_l = N_l$, then recursively reallocate the remaining bins. Note that both these strategies can result in bins for which $n_l = 0$ when $n$ is small.

\subsubsection*{Incremental sampling}

Hitherto we have assumed that sampling works by choosing a fixed batch of $n$ segments, then sending them to a rater for scoring. It is also possible to consider an interactive scenario where the rater labels segments sequentially, and the sampling procedure is refined after each new rating is received.
A convenient way to incorporate known ratings is to use them for improving the per-bin variance estimates $\hat{\sigma}_l$ in optimal allocation. We tested two ways of accomplishing this:
empirically estimate $\hat{\sigma}_l$ from the known ratings in each bin; and
learn a general mapping from metrics $\vect{y}$ to rating $x$ over \emph{all} known ratings, then use this mapping to estimate the unknown ratings in each bin, and derive 
$\hat{\sigma}_l$ from those estimates.
% These strategies are potentially complementary: for smaller $n$, metric-based estimates are likely to be advantageous, but as $n$ grows the empirical estimates will
% contain more direct information about $\sigma_l$.

\subsection{Control variates}
\label{sec:cv-methods}

The control-variates estimator makes use of an auxiliary random variable $Z$ that is standardized (has zero mean and unit variance) on the test set:
\begin{equation} \label{eqn:cvscalar}
\begin{split}
\hat{\mu} & = \bar{X}_n - \frac{\Cov(X,Z)}{\Var(Z)} \bar{Z}_n \\
          & = \bar{X}_n - \Cov(X,Z) \, \bar{Z}_n
\end{split}
\end{equation}
where $\bar{X}_n$ and $\bar{Z}_n$ are mean values over the sample, and the covariance is over the whole test set. This is the lowest-variance estimator that uses information from $Z$. It is unbiased because 
$\bar{X}_n$ is unbiased, $\Cov(X, Z)$ is independent of the current sample, and $E(\bar{Z}_n) = 0$. The control-variates estimator can be thought of as using $\bar{Z}_n$ to infer the direction that $\bar{X}_n$ has been shifted away from $\mu$ and reversing this shift by an amount that depends on the degree of correlation between $X$ and $Z$---see Figure~\ref{fig:control-variates-diagram} for an illustration. In general, $\Cov(X,Z)$ is unknown, but it can be estimated from the sample as follows:\footnote{This equation follows from expanding $\Cov(X,Z)$ over the complete test set, dropping all terms that contain the true mean of Z (0 by construction) and estimating the one term that remains from the sample. Alternatively one can choose to estimate $\Cov(X,Z)$ purely from the sample as $\sum_{i=1}^n (X_i - \bar{X})(Z_i - \bar{Z})/n$.}
\[
\Cov(X,Z) \approx \frac{1}{n} \sum_{i=1}^n X_i Z_i.
\]

The control-variates estimator can be extended to handle multiple auxiliary variables by forming a linear combination \cite{glynn2002some}:
\begin{equation} \label{eqn:cvmulti}
\hat{\mu} = \bar{X}_n - (E(\vect{Z}\vect{Z}^T)^{-1} E(X\vect{Z}))^T \,\vect{\bar{Z}}_n
\end{equation}
where $\vect{{Z}}$ is a vector of standardized variables, $\vect{\bar{Z}}_n$ is its mean over the sample, and the expectations of the covariance matrix $\vect{Z}\vect{Z}^T$ and weighted vectors $X\vect{Z}$ are taken over the test set. The latter is unknown, but as in the scalar case it can be estimated from the sample:
\[
E(X\vect{Z}) \approx \frac{1}{n} \sum_{i=1}^n X_i \vect{Z}_i.
\]

In our setting, control variates are easily derived by standardizing the metric scores $\vect{y}_i$, which are available for all segments in the test set. The resulting estimator is convenient because it is applied after sampling is complete, making it independent of the sampling method, including whether the sample is drawn incrementally or in batch mode.

\subsection{Error Bounds}
\label{sec:error}

For practical applications it is desirable to upper-bound the error $|\mu-\hat{\mu}|$ in the estimated score with some degree of confidence. Given a confidence level $\gamma$ (e.g., 0.95), we would like to find an
error bound $t$ such that:
\begin{equation} \label{eqn:bounds}
P(|\mu - \hat{\mu}| \le t) \ge \gamma
\end{equation}
A classical bound can be derived from Hoeffding's inequality, which states that equation $(\ref{eqn:bounds})$ holds if:
\[
t = R \sqrt{\frac{k_n \log(2/\delta)}{2n}},
\]
where R is the difference between the largest and smallest scores in the test set, $\delta=1-\gamma$, and $k_n = 1 - (n-1)/N$ is an adjustment for sampling without replacement 
\cite{Serfling74}. % \footnote{This factor is 1 for sampling with replacement.} 
A problem with Hoeffding's inequality is that it scales with the range of the scores and does not take variance into account, so its bound will be pessimistic if variance is small relative to the extremes. In such cases, the Bernstein bound \cite{mnih2008empirical} will be tighter:
\[
t = \hat{\sigma} \sqrt{\frac{2\log(3/\delta)}{n}} + \frac{3R\log(3/\delta)}{n},
\]
where $\hat{\sigma}$ is a sample estimate of the variance. Note that the contribution of $R$ diminishes as $1/n$ in this formula, compared with $1/\sqrt{n}$ in the Hoeffding bound. 
Both these bounds are general in the sense that they make no assumptions about the score distribution.

\section{Data}

Our development data consists of MQM ratings made available by Freitag et al. \shortcite{freitag2021experts} for 10 English-German and 10 Chinese-English ``systems'' (including human translations and MT) from the WMT 2020 news test sets \cite{barrault-etal-2020-findings}.
Each segment was annotated by three expert raters who assigned scores ranging from 0 (perfect) to 25 (worst).
There were six annotators per language pair, each of whom rated all system outputs for a set of documents comprising approximately half the complete test set (about 710 segments / rater for German, and 1000 segments / rater for Chinese). 
% Average system scores and rater assignments are shown in Appendix~\ref{sec:data}.

We created simulations for each rater and system combination, excluding the \emph{Human-A} ``system'', as it was the reference for the MT metrics we used as features. This resulted in 54 simulations for each language pair. For each simulation, the task is to predict the average score over the complete subset of segments annotated by a single rater for a single system. No knowledge of other segments, system outputs, or rater decisions is permitted to leak across simulations.
% Note that perfect performance on this task (across all raters and outputs) would be sufficient to recover the final system-level scores in Table~\ref{tab:sys-scores}, which are averaged over all raters and segments. 
As features, we used the 10 metrics submitted to the WMT 2020 metrics task \cite{mathur-etal-2020-results} that had highest 
average segment-level Pearson correlation with the MQM scores in our dev data.\footnote{We also tried using \textit{all} submitted metrics, with slightly worse results.}
These correlations are generally poor: from 0.279--0.410 for English-German, and 0.425--0.465 for Chinese-English.\footnote{For comparison, target sequence length correlations are 0.223 and 0.439 respectively (better than the three lowest-ranked metrics for Chinese).} 
% Interestingly, all of the top-performing metrics use word-embeddings in some way; this is unlike the picture at the system level \cite{freitag2021experts}, where traditional surface-based metrics like BLEU do relatively better.

To eliminate the effects of hyper-parameter tuning on the development data, we carried out  additional evaluation on a test set consisting of 
news-test data from the WMT 2021 metrics shared task \cite{freitag-etal-2021-results} for English-German (17 systems), Chinese-English (15 systems), and English-Russian (16 systems). 
This is similar to the dev data, except that only one MQM rating is available per segment. The number of rated segments was 527 for German and Russian, and 650 for Chinese. English-Russian ratings were annotated using a different MQM methodology (from Unbabel rather than Google), resulting in scores on a 0--100 scale, with 100 being best. As before, we created separate simulations for each system, omitting the human ``system'' used as a reference for the metrics. To avoid bias, rather than selecting metrics according to correlation, we chose the WMT 2021 primary submissions of two top-performing metrics from the dev data: BLEURT and COMET.\footnote{The primary submissions were \textit{BLEURT-20} and \textit{COMET-MQM\_2021}.}

Appendix~\ref{sec:data} contains further details about scores and rater assignments
for the dev and test sets.

\section{Results}

We tested the sampling and estimation strategies described in section~\ref{sec:methods} by comparing them to the baseline of simple random sampling with a mean estimator. For each simulation we considered sample sizes ranging from 5--50\% of the available data, at 5\% intervals.\footnote{Beyond 50\%, the variance of the baseline estimator becomes very low and there is limited opportunity for improvement.} For each sample size and technique for establishing $\hat{\mu}$, we drew 100 random samples, computed the average and std deviation of the error $|\mu - \hat{\mu}|$ across the samples, then averaged the results across simulations to summarize performance at that sample size. We also measured the number of ``wins''---simulations in which a technique had a lower average error than the baseline. Finally, we aggregated these results across sample sizes to summarize performance in a single number.

\subsection{Stratified sampling}

\begin{figure*}[htb]
\includegraphics[scale=0.40]{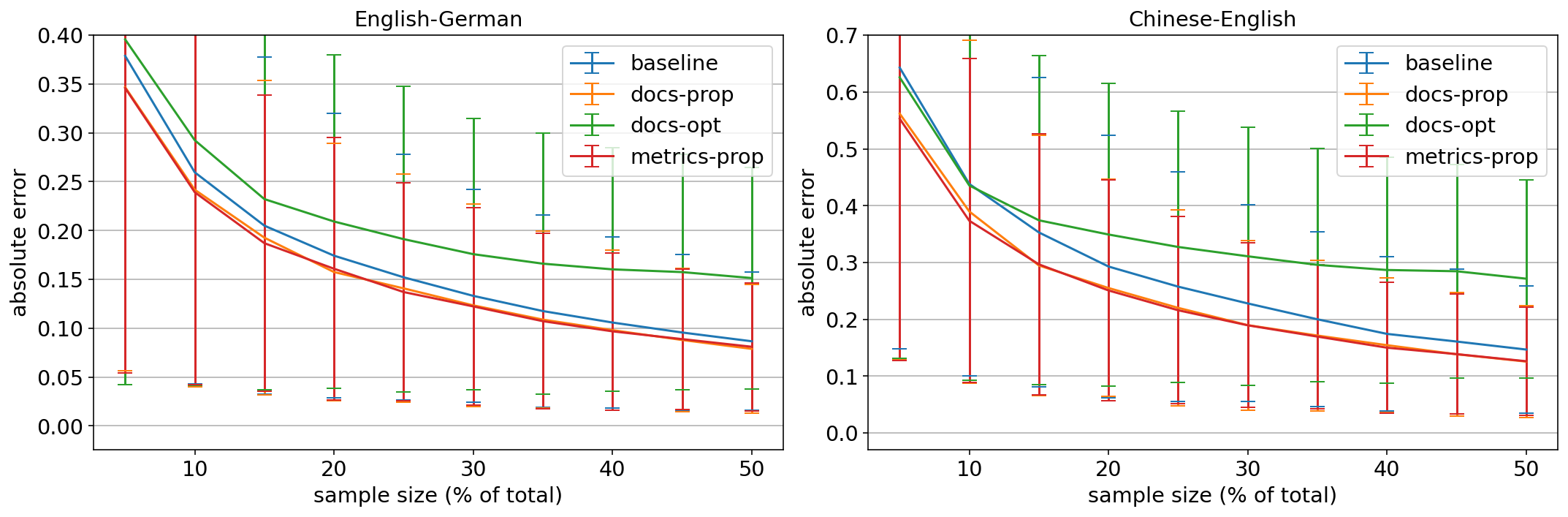}
\caption{Absolute error and standard deviation for stratified sampling methods.}
\label{fig:sampling-curves}
\end{figure*}

% \begin{figure*}[htb]
% % \includegraphics[scale=0.63]{plots/incr-sampling-curves.png}
% \includegraphics[scale=0.40]{plots/incr-sampling-curves2.png}
% \caption{Absolute error and standard deviation for incremental stratified sampling methods.}
% \label{fig:incr-sampling-curves}
% \end{figure*}

\begin{table}[bht]
\centering
\small
\begin{tabular}{ll|rrr}
\toprule
& method & \multicolumn{1}{l}{abs error} & \multicolumn{1}{l}{sdev} & \multicolumn{1}{l}{win \%} \\
\midrule
EnDe & baseline & 0.171 & 0.128 & -- \\
& docs-prop       & 0.158 & 0.118 & 75.7 \\
& docs-opt        & 0.213 & 0.145 & 32.6 \\
& metrics-prop    & 0.157 & 0.118 & 77.2 \\
\midrule
ZhEn & baseline        & 0.290 & 0.217 & -- \\
& docs-prop       & 0.250 & 0.187 & 92.4 \\
& docs-opt        & 0.356 & 0.233 & 27.2 \\
& metrics-prop    & 0.246 & 0.185 & 91.1 \\
\bottomrule 
\end{tabular}
\caption{Stratified sampling results aggregated over sample sizes from 5\%--50\%. Segment allocation is referred to as `prop' for proportional- and as `opt' for optimal-allocation with either document-based (docs) or metric-based (metrics) bin membership.
%The \emph{sdev} column contains the standard deviation of absolute errors across 100 random samples for each setting, averaged over all settings.
}
\label{tab:stratified-sampling}
\end{table}

% and depicted in figure~\ref{fig:sampling-diagram}
We begin by evaluating the stratified sampling methods described in section~\ref{sec:sampling-methods}, comparing stratification over documents and over bins defined by metric scores. The latter were formed by scoring each segment with an average of the standardized metric scores assigned to it, then sorting and partitioning so each bin contained approximately 80 segments (8x larger than the average document). More elaborate clustering and metric-selection techniques did not improve over this method.
% , nor did selecting subsets of the metrics from Table~\ref{tab:metric-corrs}. 
Performance was also quite flat as a function of bin size, though it worsened as bin size approached average document size.
We tested both stratification methods with proportional and optimal allocation using averaged metric scores as proxies for human scores when estimating the variance in each bin.

Figure~\ref{fig:sampling-curves} shows absolute error for these methods as a function of sample size, and Table~\ref{tab:stratified-sampling} summarizes aggregate performance across sizes. The general pattern is similar for both language pairs: proportional allocation with documents (\emph{docs-prop}) outperforms the random-sampling baseline; proportional allocation with metrics (\emph{metrics-prop}) behaves similarly; and optimal allocation with document bins ({\em docs-opt}) underperforms, as does optimal allocation with metric bins (not shown, as it is much worse). Optimal allocation focuses sharply on bins with high estimated variance---which will be harmful if the estimates are wrong---so we experimented with various smoothing methods, but none improved over pure proportional allocation.

Although stratification clearly reduces the error on average, the usefulness of this result is tempered by the large variances shown in Table~\ref{tab:stratified-sampling}. For any given random draw, these imply that the stratified estimate is only slightly more likely to be better than the baseline. 
Even when comparing errors averaged over 100 random draws per simulation, the stratified estimates are only better than the baseline for approximately 75\% of simulations for English-German, and 90\% for Chinese-English.

\subsubsection*{Incremental sampling}

\begin{table}[bht]
\centering
\small
\begin{tabular}{ll|rrr}
\toprule
& method & \multicolumn{1}{l}{abs error} & \multicolumn{1}{l}{sdev} & \multicolumn{1}{l}{win \%} \\
\midrule
EnDe & baseline & 0.171 & 0.128 & -- \\
& docs-incr-metrics & 0.183 & 0.132 & 44.1 \\
& docs-incr-human   & 0.231 & 0.143 & 26.7 \\
\midrule
ZhEn & baseline          & 0.290 & 0.217 & -- \\
& docs-incr-metrics & 0.346 & 0.239 & 25.4 \\
& docs-incr-human   & 0.418 & 0.251 & 27.4 \\
\bottomrule 
\end{tabular}
\caption{Incremental stratified sampling results aggregated over sample sizes from 5\%--50\%.}
% The \emph{sdev} column contains the standard deviation of absolute errors across 100 random samples for each setting, averaged over all settings.}
\label{tab:incr-stratified-sampling}
\end{table}

% Figure~\ref{fig:incr-sampling-curves} and 
Table~\ref{tab:incr-stratified-sampling} shows aggregate results for incremental stratified sampling using documents as bins, with two methods for estimating per-bin variances for optimal allocation.\footnote{We omit the corresponding curves for space reasons.} The \emph{docs-incr-metrics} method involves learning a k-nearest-neighbor (k=25) model with standardized metrics as features on all labeled segments, then using its predictions to estimate variances for the unlabeled segments in each bin. In \emph{docs-incr-human}, the variance of the segments remaining in each bin is estimated from the segments that have already been scored. Both these methods underperform the baseline; in particular, the use of a learned mapping in \emph{docs-incr-metrics} provides only modest gains over the raw averages in \emph{docs-opt}.

% This method performs similarly to \emph{docs-opt}; the use of a learned mapping yields only a slight improvement in the estimated variances compared to raw averaged metric scores, and performance remains well below the baseline. For this experiment we set $k=25$, and only updated the model at $5\%$ sampling increments (rather than after each segment). Performance was insensitive to these parameters.
% In the {\em docs-incr-human} method, the variance of the segments remaining in each bin is estimated from the segments that have already been scored; this appears to be a very poor estimator.

% The relatively low error rates in these curves for small sample sizes is an artifact of a backoff strategy in which we revert to proportional allocation for bins containing fewer than 2 scored segments. 
% In general, the variance of scored segments in a bin appears to be a very poor predictor of the variance of the remaining segments, though there is a slight improvement as the number of seen segment scores approaches $50\%$. 

\subsection{Control variates and combined results}

\begin{figure*}[htb]
\centering
\includegraphics[scale=0.40]{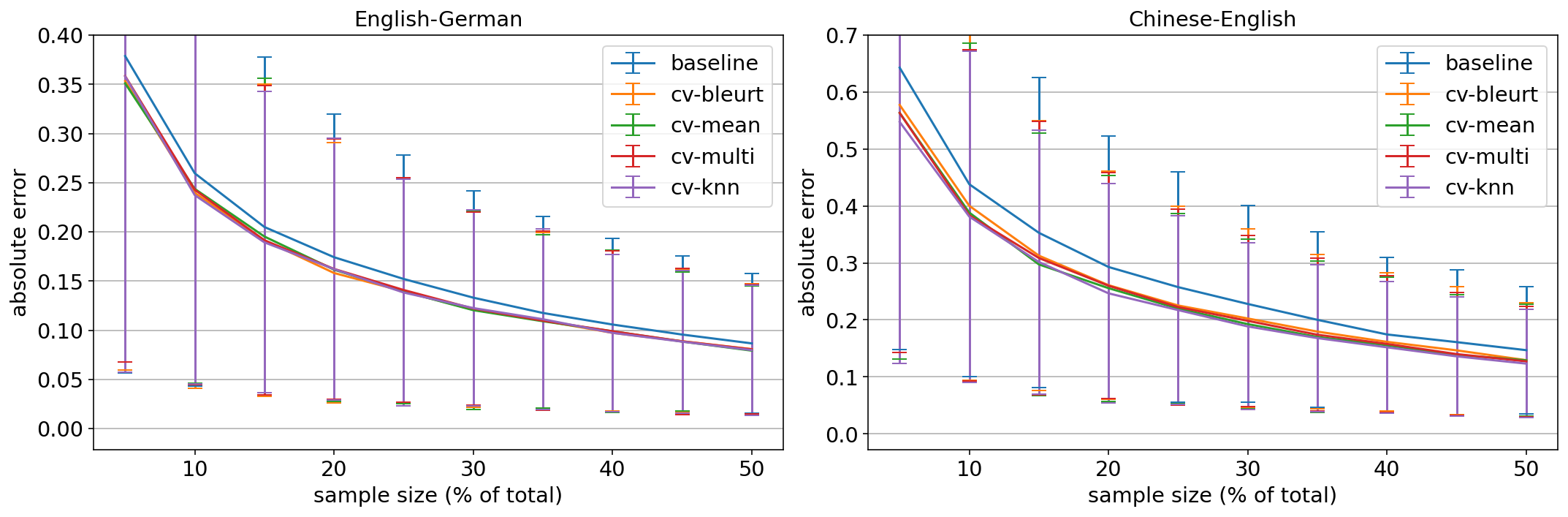}
\caption{Absolute error and std deviation for different control-variate estimators with random sampling.}
\label{fig:cv-curves}
\end{figure*}

\begin{figure*}[htb]
\includegraphics[scale=0.40]{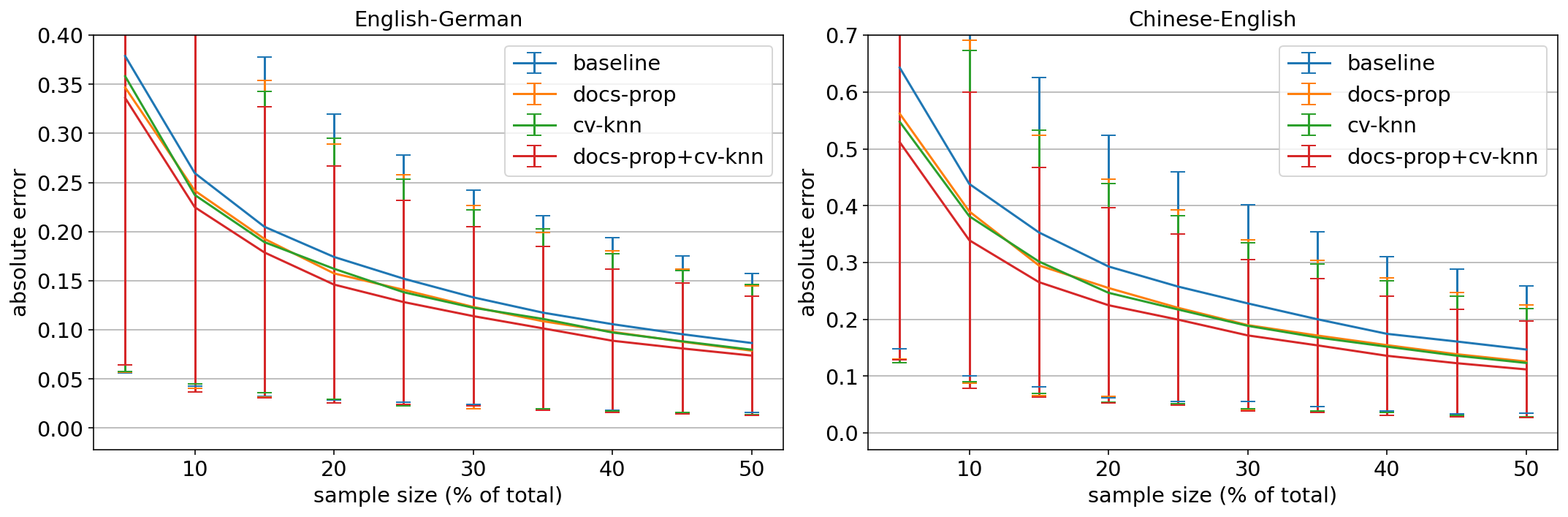}
\caption{Absolute error and std deviation for control-variate estimators and stratified sampling.}
\label{fig:comb-curves}
\end{figure*}

\begin{table}[htb]
\centering
\small
\begin{tabular}{ll|rrr}
\toprule
& method & \multicolumn{1}{l}{abs error} & \multicolumn{1}{l}{sdev} & \multicolumn{1}{l}{win \%} \\
\midrule
EnDe & baseline        & 0.171 & 0.128 & -- \\
& cv-bleurt       & 0.158 & 0.118 & 74.3 \\
& cv-mean         & 0.159 & 0.118 & 74.8 \\
& cv-multi        & 0.160 & 0.118 & 73.3 \\
& cv-knn          & 0.158 & 0.119 & 74.1 \\
\midrule
ZhEn & baseline        & 0.290 & 0.217 & -- \\
& cv-bleurt       & 0.260 & 0.193 & 84.1 \\
& cv-mean         & 0.251 & 0.188 & 88.3 \\
& cv-multi        & 0.254 & 0.188 & 88.5 \\
& cv-knn          & 0.246 & 0.185 & 92.2 \\
\bottomrule 
\end{tabular}
\caption{Control variates results aggregated over sample sizes from 5\%--50\%.}
% The \emph{sdev} column contains the standard deviation of absolute errors across 100 random samples for each setting, averaged over all settings.}
\label{tab:cv-results}
\end{table}

\begin{table}[htb]
\centering
\small
\begin{tabular}{ll|rrr}
\toprule
& method & \multicolumn{1}{l}{abs error} & \multicolumn{1}{l}{sdev} & \multicolumn{1}{l}{win \%} \\
\midrule
EnDe & baseline         & 0.171 & 0.128 & -- \\
& docs-prop        & 0.158 & 0.118 & 75.7 \\
& cv-knn           & 0.158 & 0.119 & 74.1 \\
& docs-prop+cv-knn & 0.147 & 0.110 & 88.5 \\
& metrics-prop+cv-knn & 0.156 & 0.116 & 77.8 \\
\midrule
ZhEn & baseline          & 0.290 & 0.217 & -- \\
& docs-prop         & 0.250 & 0.187 & 92.4 \\
& cv-knn            & 0.246 & 0.185 & 92.2 \\
& docs-prop+cv-knn  & 0.224 & 0.167 & 98.5 \\
& metrics-prop+cv-knn & 0.244 & 0.182 & 92.0 \\
\bottomrule 
\end{tabular}
\caption{Combined stratified sampling and control variates aggregated over sample sizes from 5\%--50\%.}
% The \emph{sdev} column contains the standard deviation of absolute errors across 100 random samples for each setting, averaged over all settings.}
\label{tab:comb-results}
\end{table}

We now turn to experiments with the control-variate estimators described in section~\ref{sec:cv-methods}. Figure~\ref{fig:cv-curves} and Table~\ref{tab:cv-results} present the results. We derived standardized scalar variates to plug into equation (\ref{eqn:cvscalar}) from: a single high-performing metric (BLEURT-extended, \emph{cv-bleurt}); the mean of all metrics (\emph{cv-mean}); and predictions from a knn model learned from all metric values on the labeled segments (\emph{cv-knn}). We also used all standardized metrics directly (\emph{cv-multi}) as input to the vector in equation (\ref{eqn:cvmulti}).\footnote{Note that the latter combines scores linearly, in contrast to the knn model.}

All tested variants give reasonable improvements over the baseline, with quite similar error rates, especially for English-German. For Chinese-English, combining all metrics with the knn model improves slightly over BLEURT-extended, reducing the absolute error by 5\%.
This may reflect somewhat higher metric correlations for this language pair.
% shown in table~\ref{tab:metric-corrs}.

%Since the error rates for control variates are remarkably similar to those for stratified sampling, a natural question is whether these techniques are somehow exploiting the same source of variance reduction despite being complementary in principle. 

As control variate estimation is applied after sampling is complete, it is straightforward to combine it with stratification. Figure~\ref{fig:comb-curves} and Table~\ref{tab:comb-results} show the results of combining
proportional stratified sampling using documents with
the best control variates estimator 
(\emph{docs-prop+cv-knn}), along with the component techniques for comparison. As one might hope, the techniques are complementary despite their similar individual performance. 
Interestingly, this is not the case when metric-based clusters are used for stratification instead of documents
(\emph{metrics-prop+cv-knn}, last line in Table~\ref{tab:comb-results}), because the same information is used for both variance-reduction techniques.
The \emph{docs-prop+cv-knn} combination produces our best results, with error reductions of 14\% and 23\% over the baseline for English-German and English-Chinese, and better average performance in almost 90\% and 100\% of simulations, respectively. Unfortunately, however, the standard deviation of these estimates remains uncomfortably close to the size of the average absolute error.

\subsection{Error estimation}

% \begin{table}[htb]
% \centering
% \small
% \begin{tabular}{r|rr|rr}
% \multicolumn{1}{c}{} & \multicolumn{2}{c}{EnDe} & \multicolumn{2}{c}{ZhEn}   \\
% \toprule
% size (\%) & Hoeff & Bern & Hoeff & Bern \\
% & slack & slack & slack & slack \\
% \midrule
% 10 & 3.61 & 5.08 &  2.88 & 4.38 \\
% 20 & 2.40 & 2.71 &  1.92 & 2.49 \\
% 30 & 1.83 & 1.89 &  1.47 & 1.81 \\
% 40 & 1.47 & 1.48 &  1.18 & 1.46 \\
% 50 & 1.20 & 1.23 &  0.96 & 1.25 \\
% \bottomrule 
% \end{tabular}
% \caption{Performance of error bounds for different sample sizes. The \emph{slack} columns give average amounts by which the bound exceeds the true error.} 
% % The \emph{under} column gives average proportion of samples for which true error was lower than the bound.}
% \label{tab:bound-results}
% \end{table}

\begin{table}[htb]
\centering
\small
\begin{tabular}{lr|rrr|rrr}
\multicolumn{2}{c}{} & \multicolumn{3}{c}{EnDe} & \multicolumn{3}{c}{ZhEn}   \\
\toprule
size &  & \multicolumn{3}{c|}{Hoeffding (4)} & \multicolumn{3}{c}{Hoeffding (7)} \\
(\%) && cal & slack & \multicolumn{1}{c}{t} & cal & slack & \multicolumn{1}{c}{t} \\
\midrule
10 & base & 92 & 0.36 & 0.61 &    89 & 0.56 & 0.90 \\
   & best & 96 & 0.40 &      &    97 & 0.49 &      \\
% \midrule
% 20 & base & 93 & 0.25 & 0.41 &    90 & 0.38 & 0.60 \\
%   & best & 96 & 0.26 &      &    96 & 0.33 &      \\
\midrule
30 & base & 93 & 0.19 & 0.31 &    90 & 0.29 & 0.46 \\
   & best & 96 & 0.20 &      &    96 & 0.25 &      \\
% \midrule
% 40 & base & 92 & 0.15 & 0.25 &    91 & 0.23 & 0.37 \\
%   & best & 96 & 0.16 &      &    97 & 0.20 &      \\
\midrule
50 & base & 92 & 0.12 & 0.20 &    90 & 0.19 & 0.30 \\
   & best & 96 & 0.13 &      &    96 & 0.16 &      \\
\bottomrule 
\end{tabular}
\caption{Performance of error bounds for different sample sizes. Statistics are averaged over simulations: \emph{cal} is \% of samples for which the true error was lower than the bound, \emph{slack} is the difference between the bound and the error, and \emph{t} is the bound. \emph{base} is the baseline estimator, and \emph{best} is \emph{docs-prop+cv-knn}.}
\label{tab:bound-compare}
\end{table}

% Despite large variance across individual samples, sampling techniques can be useful in practice if it is possible to reliably bound the error in the estimate derived from a given sample. Table~\ref{tab:bound-results} shows the performance of the error bounds
% described in section~\ref{sec:error} 
% for different sample sizes with \emph{docs-prop+cv-knn}, setting $\gamma = 0.95$.
% Both the Hoeffding and Bernstein bounds are very loose, overestimating the true error in 100\% of samples, by margins that are about an order of magnitude greater than the average error in Figure~\ref{fig:comb-curves}. We hypothesize that this is due to the large range $R$ of scores, and their highly skewed nature, with $\mu \ll R$. To test this, we recomputed the Bernstein bound with ``effective'' $R$ values of 4 and 7 for English-German and Chinese-English respectively. 
% The results (final two columns of Table~\ref{tab:bound-results}) are well calibrated 
% with much more reasonable bounds. For instance, the average bounds at 50\% sample size are
% 0.20 and 0.30 for the two language pairs. Although such ad hoc tuning is not a replicable recipe for establishing bounds in general, it does demonstrate the existence of reasonable bounds in our setting.

Despite large variance across individual samples, sampling techniques can be useful in practice if it is possible to reliably bound the error in the estimate derived from a given sample.
We computed the bounds from section~\ref{sec:error} for different sample sizes with \emph{docs-prop+cv-knn}, setting $\gamma = 0.95$.
Both the Hoeffding and Bernstein bounds are very loose, overestimating the true error in 100\% of samples, by margins that are about an order of magnitude greater than the average error in Figure~\ref{fig:comb-curves}.\footnote{Surprisingly, the Bernstein bound is somewhat worse, likely due to our small sample sizes in conjunction with the large multiplier on $R$ in the Bernstein formula.}
We hypothesize that this is due to scores having a large range $R$, and being highly skewed, with $\mu \ll R$. 

To test this, we recomputed the Hoeffding bound with empirically-determined $R$ values of 4 and 7 for English-German and Chinese-English.
As shown in Table~\ref{tab:bound-compare}, this gives results which are well calibrated (\emph{cal} > 95\%) for \emph{doc-prop+cv-knn}, with reasonable error bounds. Performance is somewhat worse for the baseline estimates, although the difference in error between the two techniques is negligible compared to the predicted bound. 
This oracle experiment suggests that it will be difficult to find non-oracle bounds that are substantially lower for \emph{doc-prop+cv-knn} than for the baseline.

\subsection{Results on test data}

\begin{figure*}[htb]
\includegraphics[scale=0.40]{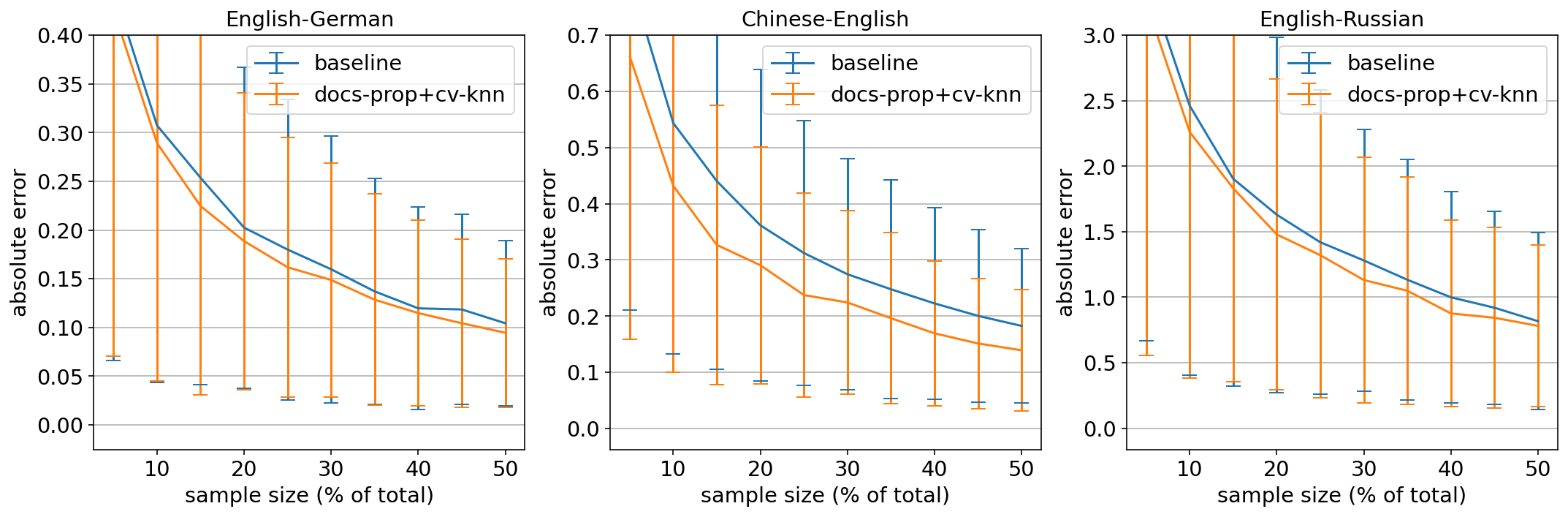}
\caption{Absolute error for control-variate estimators and stratified sampling on eval data.}
\label{fig:eval-comb-curves}
\end{figure*}

\begin{table}[htb]
\centering
\small
\begin{tabular}{ll|rrr}
& \multicolumn{1}{l}{method} & \multicolumn{1}{l}{abs err} & \multicolumn{1}{l}{sdev} & \multicolumn{1}{l}{win \%} \\
\midrule
EnDe & baseline         & 0.203 & 0.153 & -- \\
      & docs-prop+cv-knn & 0.188 & 0.140 & 78.1 \\
\midrule
ZhEn & baseline         & 0.359 & 0.267 & -- \\
      & docs-prop+cv-knn & 0.283 & 0.212 & 97.9 \\
\midrule
EnRu & baseline         & 1.601 & 1.197 & -- \\
      & docs-prop+cv-knn & 1.482 & 1.117 & 77.3 \\
\bottomrule 
\end{tabular}
\caption{Results on test data for baseline and best combined estimator aggregated over sample sizes from 5\%--50\%.}
% The \emph{sdev} column contains the standard deviation of absolute errors across 100 random samples for each setting, averaged over all settings.}
\label{tab:eval-comb-results}
\end{table}

Figure~\ref{fig:eval-comb-curves} and Table~\ref{tab:eval-comb-results} show results comparing baseline random sampling with \emph{docs-prop+cv-knn} on our evaluation set. Both the curves and the aggregate results display a similar pattern to the development results, with relatively large gains over the baseline for Chinese-English (21\% relative error reduction, wins in 98\% of simulations), and smaller ones for English-German and English-Russian\footnote{Note that the absolute errors are higher for English-Russian due to the 4x scale for ratings.} (reductions of 7\% and win rates of about 77\%). As before, standard deviations are very high.

\section{Discussion}
\label{sec:discussion}

How should we interpret these results? If we had a more reliable way of binning segments with similar human ratings, or metrics that correlated better at the segment level, it would be possible to reduce variance to levels that would permit realistic error bounds. That would enable a scenario in which we could determine the number of segments $n$ that need to be rated in order to estimate the complete test-set score to within a given tolerance. As it is, however, our error bounds are very large---and we do not manage to reduce them significantly with improved sampling and estimation methods. 
This is unlikely to change soon for complex annotation tasks like MT because
because humans are noisy raters; as shown in Table~\ref{tab:rater-accs}, they are
difficult to predict even when using other humans as oracles.
% One possibility for future work would be to tune a neural metric to learn the idiosyncrasies of a particular rater rather than using a coarse combination of multiple metrics.}

In the absence of more reliable signals for reducing variance, a way to make practical use of the techniques we study is to flip the scenario around and aim to improve the quality of an estimate made from a fixed budget of $n$ human ratings. It is common practice to obtain human annotations for only a portion of a larger test set due to time or cost constraints \cite{barrault-etal-2020-findings,freitag2021experts}. In this setting, our techniques can lead to improved estimates compared to just taking the mean of randomly-selected segments
(although there is no guarantee that they will do so for any given sample).
% Although it is hard to quantify the amount of error reduction in any particular sample, relative to just taking the mean of randomly-selected segments, our experiments have shown that our proposed techniques will lead to a reduction on average.

The risks in applying this strategy are low. Stratified sampling with proportional allocation provides an unbiased estimate of the test-set mean, with variance that is $\leq$ random sampling \cite{rice2007mathematical}, and equality only in the case that the bins have identical statistics.
The situation is trickier for control variates. In theory, the control-variate estimator is also unbiased, with lower variance than the sample mean, but this assumes that the test-set covariance $\mbox{Cov}(X,Z)$ between scores $X$ and the auxiliary variable $Z$ is known. Since we only know the scores in the sample, we must rely on an estimate for $\mbox{Cov}(X, Z)$, creating the possibility for errors if this is significantly larger than the true covariance.
However, as \citet{chaganty-etal-2018-price} point out, the error in the sample estimate for $\mbox{Cov}(X, Z)$ diminishes as $1/n$, much faster than the $1/\sqrt{n}$ rate for the error  $|\mu - \hat{\mu}|$ in the estimated score. In our data, we found no appreciable degradation of performance on small samples, even ones containing as few as 30 items.

Based on these observations, we can make the following recommendations for improving the estimated mean score of a test set containing $N$ items given a fixed number $n<N$ of items to be manually annotated:
\begin{enumerate}
    \item Use prior information such as document membership to partition items into bins, then choose items using stratified sampling as described in equation (\ref{eqn:strat-sampling}), with proportional allocation. Beware of rounding errors when only a few samples are taken from each bin.
    
    \item Use an automatic metric or other feature that correlates with human scores as a control variate in equation (\ref{eqn:cvscalar}). This step is carried out after sampling is complete, and is independent of the sampling method used. If multiple metrics are available, combine them into a single variate by averaging or applying a smooth regressor learned on the sample (knn with k=25 worked well for us). Be alert to the possibility of errors in the covariance estimate when $n$ is small ($\leq 30$).
\end{enumerate}

% If we have freedom to select the test set, what is the most effective way to use a budget of $n$ ratings? Clearly, one should select a large test set and rate only a few segments per document (or per bin established on the basis of some other prior knowledge) in order to maximize gains from stratified sampling. The situation is trickier for control variates. A larger test set will generally give a better estimate of performance in the domain of interest.  However, gains from the control-variate estimator depend on how reliably we can estimate covariances from the sample, and a faulty estimate from a sample where $n \ll N$ could cause a large skew in cases where metrics are poorly correlated and the metric test-set mean is a bad representative of the human test-set mean. We do not have access to very large test sets to experiment with, but the results for small $n$ on our small test sets show good gains over the baseline, which is reassuring. At this point, we see no reason not to use as large a test set as possible, given the usual constraints
% involved in collecting high-quality in-domain data.

% \todo{Show prediction errors for knn model trained on single metric vs all metrics?}

\section{Related Work}

\citet{chaganty-etal-2018-price} pioneered control variates for NLP evaluation, using them to improve estimates for summarization and question answering. Despite some technical differences---they measure variance ratios rather than absolute error, simulate human variance by sampling from a collection of raters, and use bootstrapped confidence intervals---their findings are roughly in line with ours. We extend their work by showing that gains from stratified sampling are complementary to those from control variates, and explore a broader range of scenarios, including using multiple variates and incremental sampling.

Recent work has investigated incremental labeling tasks and/or combining human scores with automatic metrics. 
\citet{mendoncca2021online} apply online learning algorithms to an MT system-ranking task in which different segments are selected for human evaluation on each iteration, using COMET to fill in missing human scores in WMT 2019 data. Their algorithm converges to correct results after several hundred iterations, but this condition is not detected automatically.
\citet{thorleiksdottir2021dynamic} use Hoeffding’s inequality to measure confidence in pairwise ranking decisions of varying difficulties for controlled text generation output; they consider human scores only.
\citet{singla2021using} sample foreign-language test responses for human grading, with the aim of improving over purely automatic scoring; a reverse problem to ours. 
\citet{hashimoto2019huse} propose a synergistic combination of human and automatic scoring for evaluating text generation.

Finally, there has been considerable work on measuring and rectifying inaccuracies in human annotation
\cite{sun-etal-2020-improving,wei-jia-2021-statistical,gladkoff2021measuring,paun2018comparing}.
We sidestep this issue by aiming to predict the performance of a single human rater, assuming that if this can be done accurately, conflicts among raters can be resolved in a post-processing step.

% Metrics:
% \begin{itemize}
% \item \cite{papineni2002bleu} BLEU
% \item \cite{mathur-etal-2020-results} WMT20 metrics
% \item \cite{rei-etal-2020-comet}  COMET
% \item \cite{sellam-etal-2020-bleurt}  BLEURT
% \item \cite{kocmi2021ship}  Metric eval.
% \end{itemize}

% Active learning and uncertainty:
% \begin{itemize}
% \item \cite{peris-casacuberta-2018-active} Active learning for IMT.
% \item \cite{schroder2021uncertaintybased}  Uncertainty-based active learning for NNs.
% \item \cite{tan2021diversity} Diversity Enhanced Active Learning with Strictly Proper Scoring Rules.
% \item \cite{bartolo2021models} Find hard examples interactively.
% \item \cite{xia-etal-2020-predicting}  Performance prediction for NLP.
% \item \cite{ye-etal-2021-towards}  Bounded performance prediction for NLP.
% \item \cite{glushkova-etal-2021-uncertainty-aware}  Confidence for metric scores.
% \item \cite{naeini2015obtaining}  Calibration w/ bayes binning.
% \item \cite{kang2020approximate} Precision and recall bounds for data filtering with classifiers.
% \end{itemize}

\section{Conclusion}

We investigate two classical variance-reduction techniques for improving the accuracy of sampled human ratings of MT output, measured against the mean of all ratings for a given test set. We find that stratified sampling and control variates are complementary, contributing about equally to gains of up to 20\% in average absolute error reduction compared to random sampling.
Exploiting this result to dynamically reduce annotator effort given a target error tolerance
is not feasible due to the high variance in our data,
but we propose that our techniques could instead be used to improve estimates made from a fixed annotation budget. Concrete recommendations for this scenario are provided in
section~\ref{sec:discussion}.
Our method is easy to implement, and can be applied to any setting involving averaged numerical item-wise scores where document (or other prior grouping) and automatic metric side information is available.

In future work we look forward to delving into questions raised by our results: why doesn't optimal allocation work better, particularly in the incremental setting; is there a better way to estimate variance from metrics; why aren't metric combinations more helpful; and can error bounds be improved, perhaps with bootstrapping methods?

% \section*{Acknowledgements}

\bibliography{main}
\bibliographystyle{acl_natbib}

\clearpage
\appendix

\section{Data}
\label{sec:data}

This section gives details of the development and test data used in our experiments. Table~\ref{tab:rater-work} shows the numbers of segments and documents assigned to each rater in our development data. Table~\ref{tab:sys-scores} contains the scores assigned to all ten evaluated systems; each score is an average of three rater scores per segments, averaged over all segments in the test set. Table~\ref{tab:metric-corrs} lists the selected metrics used for the development-set experiments, along with the segment-level Pearson correlation for each metric. Tables \ref{tab:rater-work-wmt21} and \ref{tab:sys-scores-wmt21} contain rater assignments and system scores for the three language pairs used in the test data.

\begin{table}[t]
\centering
\begin{tabular}{l|rr|rr}
\multicolumn{1}{l}{} & \multicolumn{2}{c}{EnDe} & \multicolumn{2}{c}{ZhEn} \\
rater & segs & docs & segs & docs \\
\midrule
rater1 & 713 & 64 & 993 & 76 \\
rater2 & 683 & 66 & 992 & 76 \\
rater3 & 705 & 66 & 1012 & 78 \\
rater4 & 709 & 65 & 996 & 79 \\
rater5 & 722 & 64 & 1021 & 77 \\
rater6 & 722 & 65 & 986 & 79 \\
\midrule
corpus & 1418 & 130 & 2000 & 155 \\
\bottomrule 
\end{tabular}
\caption{Numbers of segments and documents annotated by each rater for each system in WMT 2020 newstest.}
\label{tab:rater-work}
\end{table}

\begin{table}[t]
\centering
\begin{tabular}{lr|lr}
\multicolumn{2}{c}{EnDe} & \multicolumn{2}{c}{ZhEn} \\
system & MQM & system & MQM \\
\midrule
Human-B    & 0.75 & {\bf Human-A}   & 3.43 \\
{\bf Human-A}    & 0.91	& Human-B   & 3.62 \\
Human-P    & 1.41	& VolcTrans & 5.03 \\
Tohoku       & 2.02	& WeChat    & 5.13 \\
OPPO         & 2.25	& Tencent   & 5.19 \\
eTranslation & 2.33	& OPPO      & 5.20 \\
Tencent      & 2.35	& THUNLP    & 5.34 \\
VolcTrans    & 2.45	& DeepMind  & 5.41 \\
Online-B     & 2.48	& DiDi\_NLP  & 5.48 \\
Online-A     & 2.99	& Online-B  & 5.85 \\
\bottomrule
\end{tabular}
\caption{MQM scores for WMT 2020 outputs from \cite{freitag2021experts}. 
Scores range from 0 (perfect) to 25 (worst). The reference used for metrics is shown in bold.}
\label{tab:sys-scores}
\end{table}

\begin{table*}[t]
\centering
\begin{tabular}{lr|lr}
\multicolumn{2}{c}{EnDe} & \multicolumn{2}{c}{ZhEn} \\
metric & \multicolumn{1}{l}{$r$}  & metric & \multicolumn{1}{l}{$r$} \\
\midrule
BLEURT-extended	& 0.410 &	COMET-QE	& 0.465 \\
COMET-2R	& 0.379	& BLEURT-extended & 0.460 \\
COMET-MQM	& 0.364	& YiSi-2	& 0.453 \\
COMET-QE	& 0.358	& COMET-2R	& 0.452 \\
COMET	& 0.349	& BERT-base-L2	& 0.446 \\
COMET-HTER	& 0.326	& OpenKiwi-XLMR & 0.440 \\
OpenKiwi-XLMR	& 0.314	& BERT-large-L2	& 0.440 \\
mBERT-L2	& 0.306	& BLEURT	& 0.437 \\
prism	& 0.293	& COMET	& 0.433 \\
YiSi-1	& 0.279	& mBERT-L2	& 0.425 \\
\midrule
target-length	& 0.223	& target-length	& 0.439 \\
\bottomrule 
\end{tabular}
\caption{Segment-level Pearson correlations between selected automatic metrics and MQM ratings on system outputs from WMT 2020 newstest. The correlations shown are computed separately for each rater and system (excluding human outputs), then averaged.}
\label{tab:metric-corrs}
\end{table*}

\begin{table*}[p]
\centering
\begin{tabular}{l|rr|rr|rr}
\multicolumn{1}{l}{} & \multicolumn{2}{c}{EnDe} & \multicolumn{2}{c}{ZhEn} & \multicolumn{2}{c}{EnRu} \\
rater & segs & docs & segs & docs & segs & docs \\
\midrule
rater & 527 & 32 & 650 & 51 & 527 & 32 \\
corpus & 1002 & 68 & 1948 & 156 & 1002 & 68 \\
\bottomrule 
\end{tabular}
\caption{Numbers of segments and documents annotated by each rater for each system in WMT 2021 newstest.}
\label{tab:rater-work-wmt21}
\end{table*}

\begin{table*}[p]
\centering
\begin{tabular}{lr|lr|lr}
\multicolumn{2}{c}{EnDe} & \multicolumn{2}{c}{ZhEn} & \multicolumn{2}{c}{EnRu}\\
system & MQM & system & MQM & system & MQM \\
\midrule
{\bf ref-C}	        & 0.51	& {\bf ref-B}	        & 4.27	& {\bf ref-A}	        & 99.65 \\
ref-D	        & 0.52	& ref-A	        & 4.35	& ref-B	        & 98.40 \\
ref-B	        & 0.80	& metricsystem1	& 4.42	& Facebook-AI	& 92.75 \\
VolcTrans-GLAT	& 1.04	& metricsystem4	& 4.62	& Online-W	    & 91.80 \\
Facebook-AI	    & 1.05	& NiuTrans	    & 4.63	& metricsystem4	& 91.25 \\
ref-A	        & 1.22	& SMU	        & 4.84	& metricsystem5	& 90.88 \\
Nemo	        & 1.34	& MiSS	        & 4.93	& metricsystem1	& 90.79 \\
HuaweiTSC	    & 1.38	& Borderline	& 4.94	& metricsystem2	& 89.86 \\
Online-W	    & 1.46	& metricsystem2	& 5.04	& Online-A	    & 87.87 \\
UEdin	        & 1.51	& DIDI-NLP	    & 5.09	& Nemo	        & 87.50 \\
eTranslation	& 1.70	& IIE-MT	    & 5.14	& Online-G	    & 87.22 \\
VolcTrans-AT	& 1.74	& Facebook-AI	& 5.21	& Manifold	    & 86.86 \\
metricsystem4	& 2.05	& metricsystem3	& 5.39	& Online-B	    & 85.66 \\
metricsystem1	& 2.07	& Online-W	    & 5.57	& metricsystem3	& 85.65 \\
metricsystem3	& 2.27	& metricsystem5	& 6.39	& NiuTrans	    & 83.47 \\
metricsystem2	& 2.58	&               &       & Online-Y	    & 79.27 \\
metricsystem5	& 2.61	&&&& \\			
\bottomrule
\end{tabular}
\caption{MQM scores for WMT 2021 outputs from \cite{freitag-etal-2021-results}. 
Scores range from 0 (perfect) to 25 (worst), except for English-Russian, where they range from 0 (worst) to 100 (perfect).
The reference used for metrics is shown in bold.}
\label{tab:sys-scores-wmt21}
\end{table*}

\clearpage

\section{Variabililty in human scores}
\label{sec:variability}

A difficulty in predicting human ratings is that humans are noisy annotators
\cite{wei-jia-2021-statistical}. To quantify the noise in our data, 
we computed the error when predicting each rater's average score over their assigned segments using the average of the other two raters who also rated those segments.
Table~\ref{tab:rater-accs} shows that this varies substantially across raters and languages, with the hardest-to-predict rater's error being over 3x that of the easiest-to-predict rater in both languages, and Chinese-English errors being higher than English-German. (Variance across raters may be due in part to differences in their assigned subsets of segments, as some segments are harder to rate than others. Variances across languages is likely due to Chinese-English system scores being higher (worse) than German-English scores.)
Comparing the average errors of 0.3 and 0.8 for English-German and Chinese-English to Figure~\ref{fig:comb-curves}, we observe that only a small number of samples (less than 10\%) of a particular annotator's own ratings are sufficient to predict their test-set score with greater precision than knowing the average of other raters' scores over the whole test set (a rough proxy for the ``true'' test-set score).

\begin{figure*}[!htb]
\centering
\includegraphics[scale=0.63]{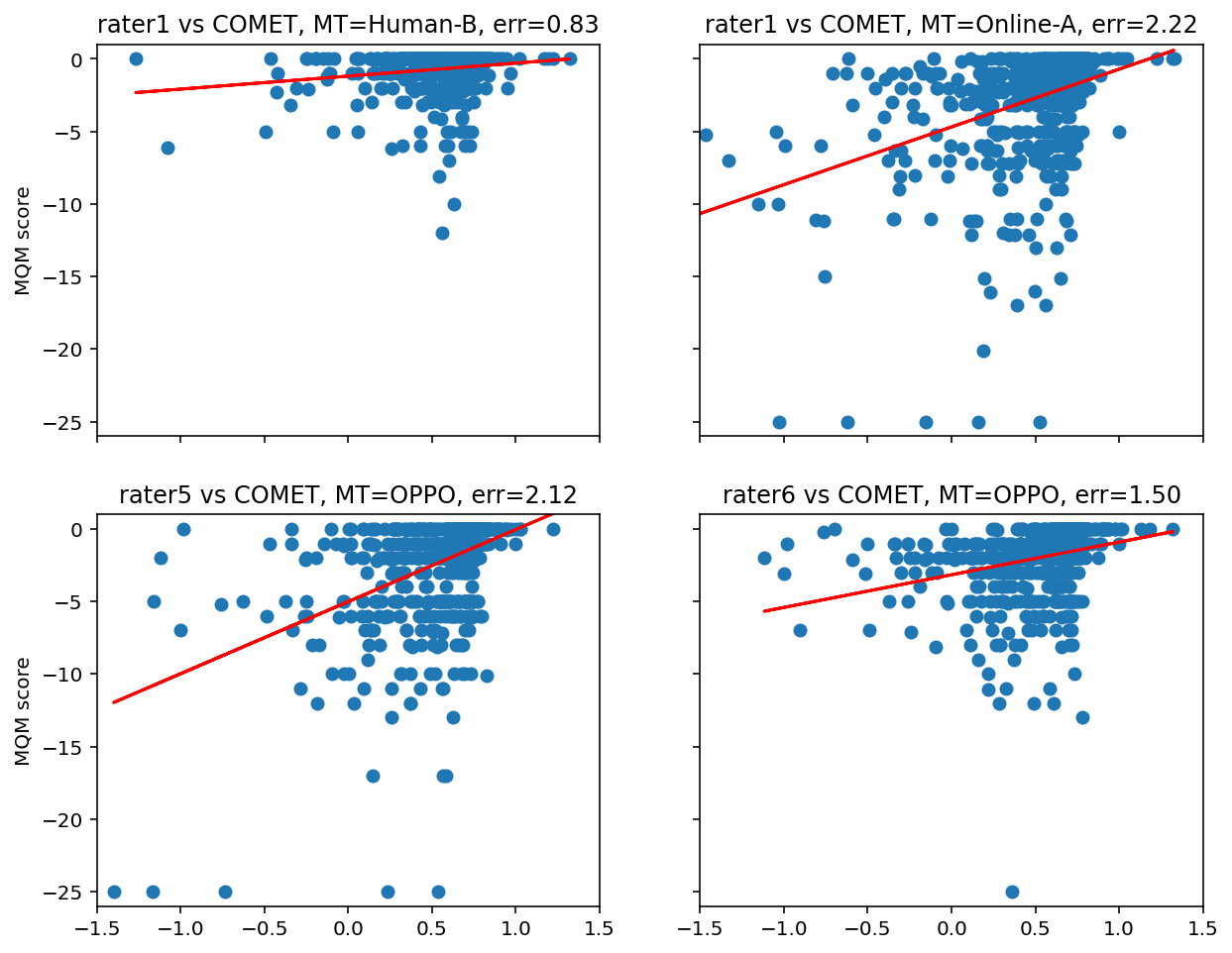}
\caption{Example WMT20 EnDe human MQM versus COMET scores for the same rater but different MT systems (top panels), and different raters but the same MT system (bottom panels). Each point represents a single segment, and the lines show the best linear fit. Errors are average absolute segment-level differences between the line and the points.}
\label{fig:metric-calibration}
\end{figure*}

A key element of our technique is using automatic MT metrics to predict human scores at the segment level. Figure~\ref{fig:metric-calibration} shows scatter plots for a single high-performing metric (COMET) that illustrate the challenges with this: the relation with MQM scores is noisy and non-linear, and there are extreme outliers due to segments that were assigned the worst possible MQM score. Furthermore, as indicated by the slope of the regression lines, the relation can vary substantially across different settings, even for different systems scored by a single rater, or for the same system scored by different raters. This implies that a strategy of pre-calibrating a particular metric on data that is independent of the current rater and system is likely to be ineffective for our problem.

\begin{table}[t]
\centering
\begin{tabular}{l|r|r}
\multicolumn{1}{l}{} & EnDe & ZhEn  \\
\midrule
rater1 & 0.13 & 0.37 \\
rater2 & 0.22 & 0.60 \\
rater3 & 0.47 & 0.38 \\
rater4 & 0.32 & 1.55 \\
rater5 & 0.14 & 1.40 \\
rater6 & 0.33 & 0.69 \\
\midrule
avgs & 0.27 & 0.83 \\
\bottomrule 
\end{tabular}
\caption{Absolute errors when predicting each rater's score from the average of other raters' scores. Numbers shown are averages over all systems and all segments annotated by the given rater.}
\label{tab:rater-accs}
\end{table}

\end{document}